\documentclass{article}
\usepackage[T1]{fontenc}
\usepackage[utf8]{inputenc}
\usepackage{lmodern}
\usepackage{libertine}
\usepackage{microtype}
\usepackage{sourcecodepro}

\usepackage{bib}
\usepackage{macros}
\usepackage{layout}
\newcommand{\mup}{$\mu$P}
\newcommand{\gmult}{g}
\newif\ifarxiv
\arxivtrue

\hypersetup{pdftitle={Lazy and rich training regimes}}

\title{The lazy (NTK) and rich (\mup) regimes:\\ A gentle tutorial}
\author{
Dhruva Karkada\thanks{\ttfamily{dkarkada@berkeley.edu}} \\ UC Berkeley}

\begin{document}

\maketitle
\thispagestyle{empty}
 
\begin{abstract}

A central theme of the modern machine learning paradigm is that larger neural networks achieve better performance on a variety of metrics. Theoretical analyses of these overparameterized models have recently centered around studying very wide neural networks. In this tutorial, we provide a nonrigorous but illustrative derivation of the following fact: in order to train wide networks effectively, there is only one degree of freedom in choosing hyperparameters such as the learning rate and the size of the initial weights. This degree of freedom controls the \textit{richness} of training behavior: at minimum, the wide network trains lazily like a kernel machine, and at maximum, it exhibits feature learning in the active \mup\ regime. In this paper, we explain this richness scale, synthesize recent research results into a coherent whole, offer new perspectives and intuitions, and provide empirical evidence supporting our claims. In doing so, we hope to encourage further study of the richness scale, as it may be key to developing a scientific theory of feature learning in practical deep neural networks.
\end{abstract}

\vspace{-1em}

\renewcommand{\contentsname}{\vspace{-12pt}}
\tableofcontents
\newpage

\section{Introduction}
\label{sec:intro}

The modern era of machine learning is characterized primarily by the use of large models. In practice, these models are both deep (i.e., consisting of many layers transforming the data in series) and wide (i.e., having hidden representations with higher dimension than the data itself.) For example, the most performant WideResNet learning CIFAR10 consists of 28 layers, and just the second hidden feature map has more than 50 times as many features as the original input \parencite{zagoruyko2016wide}. Such a long series of high-dimensional transformations can induce undesirable behaviors that have no low-dimensional analog, and we must be careful to avoid them.

To understand this, consider that training deep networks consists of alternating two complementary processes: a feedforward inference and a backpropagating update. We want to ensure that these processes remain well-behaved throughout training: feedforward outputs should evolve appreciably towards the labels in finite time, and backpropagation should induce updates in the hidden representations that allow optimization to proceed stably without stalling or exploding. How might we choose our model hyperparameters to ensure these two desiderata? This tutorial aims to transparently answer this question.

\begin{figure}[t]
  \includegraphics[width=\textwidth]{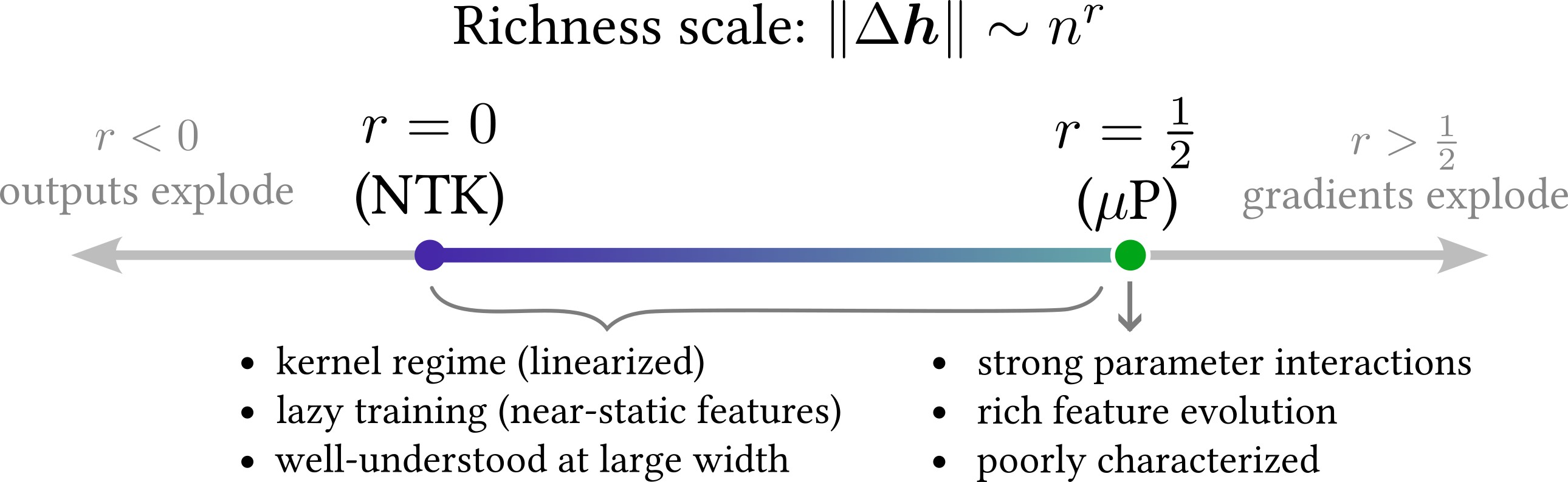}
  \caption{For well-behaved sufficiently-wide models, training behavior is characterized by a single \textit{richness} hyperparameter $r$ prescribing how the size of the hidden representation updates $\norm{\Delta \vh}$ scales with model width $n$. At finite width, model behavior changes smoothly between the NTK endpoint $r=0$ and the \mup\ endpoint $r=1/2$, but in the thermodynamic limit ($n\to\infty$) there is a discontinuous phase transition separating active \mup\ behavior from lazy $r<1/2$ behavior.}
  \label{fig:richness}
\end{figure}

We will find that codifying these desiderata as quantitative training criteria naturally constrains our choice of hyperparameters. Furthermore, it is not too difficult to solve this constraint problem, and in the end we will be left with only one degree of freedom: the size of the updates to the hidden representations. If we choose the updates to be small, we recover the lazy kernel regime, in which the model is linearized in its parameters throughout training and the hidden representations change negligibly. If the updates are large, we recover the rich (or active) \mup\ limit, in which the model learns nontrivial (and often better-generalizing) features. We depict this structure in \cref{fig:richness}.

We emphasize that this one-dimensional \textit{richness scale} follows naturally and uniquely from enforcing our training criteria. In addition, though we analyze a toy model (namely, a 3-layer linear network), our derivation faithfully captures the essence of the rigorous proofs in the literature. On the way, we will uncover and justify several key intuitions about training wide networks with gradient descent:
\begin{enumerate}
    \item Achieving stable, nontrivial, and maximal training requires controlling the relative sizes of the backpropagating gradient and the feedforward signal.
    \item Lazy training corresponds to a linearized kernel regime; representation learning (i.e., non-negligible feature evolution) necessarily does not.
    \item Weights update to align with learned representations.
    \item Small outputs at initialization are necessary to achieve rich feature-learning behavior.
\end{enumerate}

The richness scale offers a useful probe that both theorists and empiricists may tune in order to study the phase transition between the lazy kernel regime and the realistic feature-learning regime. In particular, it may enable systematic study of the performance gap between NTK learners and practical neural networks. (After all, wide networks in the rich regime are likely to describe practical neural networks more aptly than the NTK precisely because they learn features even in the infinite-width limit.) In fact, understanding the rich regime as well as we understand the kernel regime is arguably \textit{the} major open problem in developing a scientific theory of practical deep learning. Our hope here is to elucidate the ideas underlying this line of research.

\newpage

\subsection{Related work}
\label{sec:related_work}

The main results presented in this work are stated and proved rigorously in \textcite{yang2021tensor}. We show here that simple scaling arguments can be used to arrive at the same conclusions nonrigorously. What we call the \textit{richness scale} is denoted $\text{UP}_r$ in their Figure 5, although our richness hyperparameter $r$ differs from their scaling exponent $r$ in polarity (i.e., they define $r$ such that $r=0$ is \mup\ and $r=1/2$ is NTK). In addition, they relax our maximality criterion and allow both layerwise learning rates and gradient multipliers, arriving at a much larger space of possible hyperparameter scalings. However, as we argue in \cref{ssec:roadmap} and \cref{sssec:gradmults}, we don't win much by considering this general case, so we restrict our attention for conceptual clarity.

This tutorial borrows from and extends the derivation presented in \textcite{yang2023spectral}, which emphasizes controlling the spectral norm of weight matrices in order to achieve \mup\ training. In contrast, we avoid referencing particular matrix norms, instead focusing our attention on the size of the representations and gradients; this allows us to treat both feature learning and lazy training. As a result, we disambiguate the criteria that are necessary for well-behaved training from those that yield feature-learning behavior. In addition, our framework shows that \mup\ is \textit{necessary} for maximal, stable, active training, while they focus on demonstrating sufficiency. 

The lazy NTK regime was proposed by \textcite{jacot2018neural} and has since been well-studied. For example, \textcite{arora2019exact} and \textcite{yang2020tensor} provide the NTK for architectures beyond MLPs; works including \textcite{allen2019convergence} and \textcite{du2019gradient} provide convergence guarantees; and works such as \textcite{bordelon2020spectrum} and \textcite{simon2021eigenlearning} provide closed-form estimates of generalization error in terms of the NTK eigensystem. On the other hand, works such as \textcite{chizat2019lazy}, \textcite{geiger2020disentangling}, and \textcite{bordelon2023self} analyzed wide models away from the lazy regime using a ``model rescaling'' approach. In \cref{sssec:rescaling} we demonstrate that our framework accommodates this method of achieving active training, although we argue that model rescaling can induce pathological behavior if used rashly. We demonstrate this by using our framework to reveal the underlying cause of the undesired behaviors reported in those works.

Rigorous proofs regarding the linearization of models with NTK parameterization were provided in \textcite{lee2019wide}, \textcite{huang2020dynamics}, and \textcite{liu2020linearity}. To our knowledge, the nonrigorous derivation we provide in \cref{apdx:linearization} is the first to estimate the change in the tangent kernel as a function of training richness.

The experiments we perform to provide empirical evidence are inspired by those in \cite{yang2023spectral}, \cite{vyas2024feature}, and \cite{shan2021theory}. 

\subsection{Mathematical notation}
\label{ssec:notation}

We use lowercase boldface for vectors and uppercase boldface for matrices. We use $\norm{\va}$ for the Euclidean vector norm, $\transpose{\va}\vb$ for the Euclidean inner product, and $\va\otimes\vb$ for the outer product. We use subscripts to index the layers and parenthesized superscripts to denote tensor indices (e.g., $\mW_\ell^{(ij)}$ refers to the $i,j$ component of the $\ell^\text{th}$ weight matrix). We use the Einstein summation convention, under which repeated tensor indices denote a sum along that axis (e.g., $\va^{(i)}\vb^{(i)}=\transpose{\va}\vb$ but $\va^{(i)}\vb^{(j)}=\va\otimes\vb$). We use $\Delta$ to denote changes across a single optimization step (often, between initialization and the first update). We say a matrix $\mM$ is ``aligned'' with a vector $\vv$ if $\norm{\mM\hat\vv}$ typically dominates $\norm{\mM\hat\vu}$ for some random isotropic unit vector $\hat\vu$.

Throughout this tutorial, we consider how certain scalars asymptotically scale with the network width. For conciseness, we use binary relation symbols instead of Bachmann-Landau (big-O) notation to denote asymptotic scalings. Specifically, we use $\sim$ to mean ``scales with'', i.e., $a\sim b$ means $a=\Theta(b)$; $a\gtrsim b$ means $a=\Omega(b)$; $a\lesssim b$ means $a=\mathcal{O}(b)$; $a \gg b$ means $a=\omega(b)$; and $a \ll b$ means $a=o(b)$. 
\section{The richness scale}
\label{sec:derivation}

\subsection{Roadmap}
\label{ssec:roadmap}

We are interested in training wide neural networks on a supervised learning task. To study this, we will analyze a 3-layer linear model. This toy model is realistic enough to capture the salient aspects of signal propagation in practical neural networks, yet is simple enough that theoretical analysis is straightforward. The arguments we will use in our analysis faithfully capture the essence of the rigorous proofs in the literature. Note however that these signal propagation arguments are rooted in the behavior of the model near initialization, and we do not attempt to conclude anything about convergence rates or the solution quality. We provide empirical evidence that the conclusions we draw are applicable to practical networks in \cref{sssec:realisticmodels}.

As the network width becomes large, it becomes important to control the size of the gradient relative to the size of the feedforward representations. Anticipating this, we will factor the weight matrices into a fixed coefficient $\gmult$ (a scalar \textit{gradient multiplier}) and a trainable part $\mW$ (the remaining matrix). In \cref{sssec:gradmults}, we show that this approach is equivalent to simply assigning each weight matrix its own learning rate.\footnote{Since layerwise learning rates introduce a redundancy, the larger hyperparameter configuration space can still be divided into equivalence classes, yielding multiple parameterizations which behave identically throughout training. We connect our exposition to the various parameterizations in the literature in \cref{sssec:gradmults}.}

\begin{figure}[h!]
  \includegraphics[width=\textwidth]{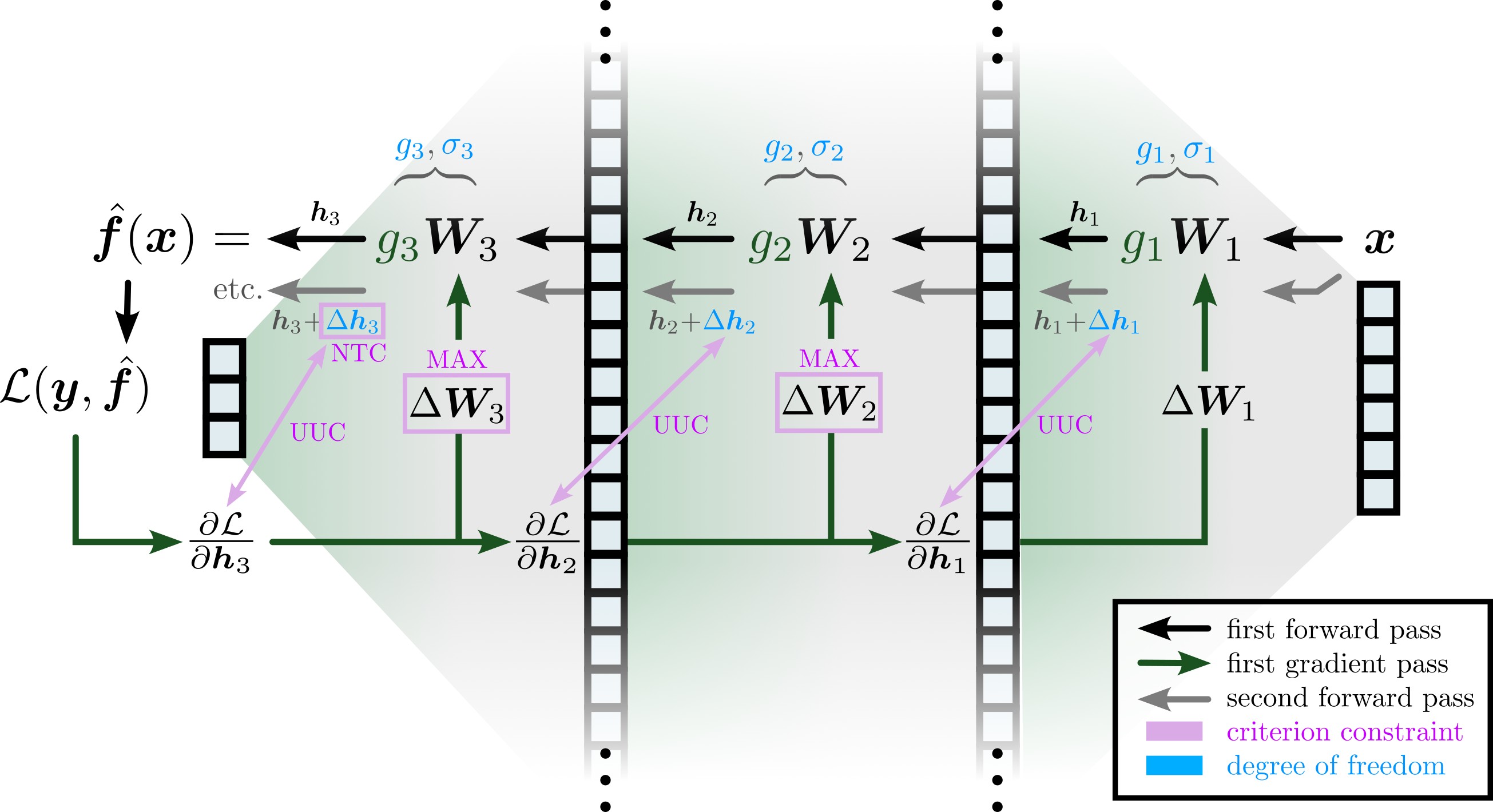}
  \caption{Signal propagation diagram for our wide 3-layer linear model. This diagram visually depicts the logical flow of our main derivation: we analyze the first forward pass, the first backward pass, and the second forward pass to enforce our training criteria (shown in pink) and constrain our initial nine degrees of freedom (shown in blue). We depict the forward pass signals flowing right to left to match the convention for matrix multiplication.
  }
  \label{fig:derivation}
\end{figure}

\newpage
\begin{mainbox}{3-layer linear model.}
    We will train the model
    \[\vh_3(\vx)\defn\gmult_3\mW_3\gmult_2\mW_2\gmult_1\mW_1\vx\]
    by gradient descent with constant $\Theta(1)$ learning rate on a standard loss function $\loss(\vy, \vh_3(\vx))$. We will use $\ell$ to index the layers, $1 \leq \ell \leq 3$. See \cref{fig:derivation}.
    \par
    The $\mW_\ell$ are weight matrices whose elements are the trainable parameters. We use Gaussian initialization with scale $\sigma_\ell$ (i.e., $\mW_\ell^{(ij)}$ drawn from $\mathcal{N}(0, \sigma_\ell^2)$). Each $\mW_\ell$ pairs with a fixed gradient multiplier $\gmult_\ell$: increasing $\gmult_\ell$ while keeping $\gmult_\ell\sigma_\ell$ fixed increases the size of the gradient received by $\mW_\ell$ while maintaining the size of the feedforward signal.
    
    For convenience, we additionally define the $\ell^\text{th}$-layer \textit{representations} recursively as
    \begin{equation}
        \label{eq:ff}
        \vh_\ell(\vx) \defn \gmult_\ell\mW_\ell \vh_{\ell-1}(\vx) \quad\qtext{with base case} \vh_0(\vx)=\vx.
    \end{equation}
    For much of our analysis we consider a single fixed $\vx$, so we abbreviate $\vh_\ell(\vx)$ as $\vh_\ell$. Define
    \[n_\ell \defn \dim \vh_\ell\]
    to be the dimension of a representation. The learning task specifies the input and output dimensions, $n_0$ and $n_3$. We will take a wide network limit with a single width scale, $n \sim n_1 \sim n_2 \gg n_0 \sim n_3 \sim 1$, but for clarity we will distinguish the $n_\ell$ in our derivations.

    Now we can justify needing three layers: as we'll later see, the signal propagation properties of a layer depend on the layer's shape ($\mathtt{shape}= \mathtt{fanin}/\mathtt{fanout}$), and our model can treat the three possibilities $\mathtt{shape}\ll1$, $\mathtt{shape}\sim 1$, and $\mathtt{shape}\gg1$.
\end{mainbox}

\cref{eq:ff} prescribes the feedforward signal on some chosen input $\vx$. How does this signal change after a gradient descent step? After updating the weights by $\Delta\mW_\ell$, the new representations are
\[\vh_\ell+\Delta\vh_\ell=\gmult_\ell\left((\mW_\ell+\Delta\mW_\ell)(\vh_{\ell-1}+\Delta\vh_{\ell-1})\right).\]
Subtracting the original representation, we find that the representation update is
\begin{equation}
    \label{eq:update}
    \Delta\vh_\ell = 
    \underbrace{\gmult_\ell\Delta\mW_\ell\vh_{\ell-1}}_{\text{layer}} + \underbrace{\gmult_\ell\mW_\ell\Delta\vh_{\ell-1}}_{\text{passthrough}} + \underbrace{\gmult_\ell\Delta\mW_\ell\Delta\vh_{\ell-1}}_{\text{interaction}}.
\end{equation}
For stable, well-behaved training, we would like to control these updates. In lieu of this, let us understand the terms involved. The \textit{layer contribution} is induced by the update to the current layer's weights. The \textit{passthrough contribution} is induced by the update to the previous representation passing through the old weights. The \textit{interaction contribution} captures the interaction between the layer update and the previous representation update.

We now posit that training will proceed stably and quickly if our model satisfies three key criteria. These criteria underpin our derivation of the training regimes, so it's worth understanding and remembering them. 

\newpage

\begin{mainbox}{Criteria for well-behaved training.}
\begin{enumerate}
    \item \textbf{Nontriviality Criterion.}
    \begin{equation*}
        \label[criterion]{crit:ntc}\tag{NTC}
        \norm{\Delta \vh_3} \sim 1
    \end{equation*}
    The updates to the outputs shouldn't scale with the width. This ensures that the loss decreases at a width-independent rate.
    
    \item \textbf{Useful Update Criterion.}
    \begin{equation*}
        \label[criterion]{crit:uuc}\tag{UUC}
        \left|\transpose{\pdv{\loss}{\vh_\ell}}\Delta \vh_\ell\right| \sim 1 \qquad \text{for } \ell\geq 1
    \end{equation*}
    Each representation update should contribute to optimizing the loss.

    \item \textbf{Maximality Criterion.}
    \begin{equation*}
        \label[criterion]{crit:max}\tag{MAX}
        \norm{\gmult_\ell\Delta \mW_\ell \vh_{\ell-1}} \sim \norm{\Delta \vh_\ell}
    \end{equation*}
    A layer's weight update should contribute non-negligibly to the following representation update (i.e., the layer contribution should not be dominated, c.f. \cref{eq:update}).
\end{enumerate}
\end{mainbox}
\newpage

The \cref{crit:ntc} is a single constraint that applies to the last layer. The \cref{crit:uuc} provides one constraint per layer, and is instrumental in guaranteeing that the gradients are appropriately-sized. The \cref{crit:max} provides one constraint per layer, although it is trivially satisfied in the read-in layer since the inputs are fixed. Although the \cref{crit:max} is not strictly necessary for stable training, it is desirable nonetheless since it prevents ``dud layers'' (i.e., effectively frozen layers.)

Taken together, these provide six constraints for our 3-layer model. Let us now count the number of degrees of freedom we have in our initialization scheme. We are free to choose the six hyperparameters $\gmult_\ell$ and $\sigma_\ell$. It turns out that we will have a choice in the size of the representation updates, $\norm{\Delta\vh_\ell}$. (The initial representation sizes, $\norm{\vh_{\ell}}$, are determined by our choice of $\gmult_\ell$ and $\sigma_\ell$.) This totals nine degrees of freedom. Satisfying the six constraints leaves three remaining degrees of freedom. We will use two of those degrees of freedom to fix the initial $\norm{\vh_2}$ and $\norm{\vh_1}$ so that the activations are $\Theta(1)$. The remaining degree of freedom will control the richness of training, i.e., the degree to which the model trains in the kernel regime or the feature learning regime.\footnote{It is not hard to extend this constraint-counting argument to models of arbitrary depth.}

To resolve this constraint problem, we will consider training our model on a single training sample. Then, we will manually perform forward and backward passes until we can apply all our criteria to solve for the hyperparameters. We visually depict the logical structure of our derivation in \cref{fig:derivation}.

\subsection{Deriving the richness scale}
\label{ssec:solving}

We begin by using the feedforward equations to enforce that the hidden activations $\vh_\ell^\idx{i}$ are initially $\Theta(1)$, i.e., $\norm{\vh_\ell}^2 \sim n_\ell$ for $\ell\in{1,2}$ (assuming the input also satisfies $\norm{\vh_0}^2 \sim n_0$).

Note that we do not have the freedom to choose the output activations to be $\Theta(1)$ without relinquishing the freedom to choose between lazy and rich training. This is because the output layer must satisfy the \cref{crit:ntc}, an extra constraint from which the hidden layers are free. We do know that the output activations shouldn't blow up with width, or else the size of the error signal $\norm{\vy-\vh_3}$ will also blow up. But the network outputs can be arbitrarily small in principle; near-zero outputs don't change the size of the error signal.

With these in mind, we take the squared vector norm of \cref{eq:ff}, apply a central-limit-style argument to the right hand side, and enforce our desired activation norm, finding

\begin{equation}
    \label{eq:ffsize}
    \gmult_\ell^2 \sigma_\ell^2 n_{\ell-1} \setsim 1 \qtext{for} \ell\in{1,2} \qqtext{and} \gmult_3^2 \sigma_3^2 n_{2} \lesssim 1.
\end{equation}

Our first forward pass has yielded three useful relations for our hyperparameters. Now let's calculate what happens during the first backwards pass. In \cref{sssec:weightalign} we show that gradient descent prescribes that the weights update to align with their inputs:

\begin{equation}
    \label{eq:weightupdate}
    \Delta \mW_\ell = -\gmult_\ell \pdv{\loss}{\vh_\ell} \otimes \vh_{\ell-1}
\end{equation}

where we adopt the ``denominator layout'' for gradients (i.e., $\pdv*{\loss}{\vh_\ell}$ is a column vector).

This derivative of the loss with respect to the representations will play a key role in our analysis, so let's use \cref{eq:ff} to find a recursive expression for it:

\begin{equation}
    \label{eq:repgradient}
    \pdv{\loss}{\vh_{\ell-1}} = \gmult_\ell  \transpose{\mW_\ell} \pdv{\loss}{\vh_{\ell}}
    \qqtext{which implies} \norm{\pdv{\loss}{\vh_{\ell-1}}}^2 \sim \gmult_\ell^2 \sigma_\ell^2 n_{\ell-1} \norm{\pdv{\loss}{\vh_{\ell}}}^2
\end{equation}

where the norm relation follows from applying the central-limit-style argument and using the fact that the factors in the right hand side are uncorrelated.

With these two backprop equations, we can already proceed to the second forward pass and evaluate the representation update given by \cref{eq:update}:

\begin{align}
    \Delta\vh_\ell 
    &= \underbrace{-\gmult_\ell^2 \norm{\vh_{\ell-1}}^2 \pdv{\loss}{\vh_\ell}}_{\text{layer}} \underbrace{-\gmult_\ell^2 \gmult_{\ell-1}^2 \norm{\vh_{\ell-2}}^2 \mW_\ell\transpose{\mW_\ell} \pdv{\loss}{\vh_\ell}}_{\text{passthrough (layer term)}}
    + \underbrace{\cdots}_{\text{passthrough (other terms)}} + \underbrace{\cdots}_{\text{interaction}} \nonumber\\
    &\approx -\gmult_\ell^2 \left(\norm{\vh_{\ell-1}}^2 + \gmult_{\ell-1}^2 \norm{\vh_{\ell-2}}^2 \mW_\ell\transpose{\mW_\ell}\right) \pdv{\loss}{\vh_\ell} 
    \label{eq:update-solved}
\end{align}

where, in evaluating the passthrough, we kept the first term in $\Delta\vh_{\ell-1}$ and used \cref{eq:repgradient}. With some prescience, we've ignored terms that contribute negligibly. (One can guess that these terms have negligible contribution because the factors within each don't align with each other.)

Let's stop and make some important observations. The first term in \cref{eq:update-solved} is the layer contribution, and the second term is the non-negligible part of the passthrough contribution. By the \cref{crit:max}, we know the first term cannot be dominated by the second. This tells us that $\Delta\vh_\ell$ has a large component in the direction of $\pdv*{\loss}{\vh_\ell}$ -- the representation update is aligned with the loss gradient with respect to the representation. Note that gradient descent prescribes exactly this kind of alignment in the weights; the \cref{crit:max} ensures that the representations enjoy this same alignment, inducing a satisfying duality between the weights and the representations. This retroactively provides further justification for the \cref{crit:max}.

We can exploit this alignment to express the \cref{crit:uuc} in a friendlier way:
\begin{equation}
    \label{eq:uuc-aligned}
    \norm{\pdv{\loss}{\vh_\ell}}\norm{\Delta\vh_{\ell}}\setsim 1 \qqtext{for} \ell\geq 1
\end{equation}
which, applied to \cref{eq:repgradient}, gives us the constraint relation
\begin{equation}
    \label{eq:sigma-constrained}
    \norm{\Delta\vh_{\ell}}^2 \setsim \gmult_\ell^2 \sigma_\ell^2 n_{\ell-1} \norm{\Delta\vh_{\ell-1}}^2 \qqtext{for} \ell\geq 2.
\end{equation}

But the feedforward constraint \cref{eq:ffsize} also applies to layer 2. This gives us our first result:
\begin{equation}
    \label{eq:sameupdate}
    \norm{\Delta \vh_1} \sim \norm{\Delta \vh_2} \defn \norm{\Delta \vh}.
\end{equation}
We've learned that the wide hidden representations all evolve at the same rate. We can drop the layer subscripts and use $\norm{\Delta \vh}$ to refer to the scale of the hidden representation updates.

Now, let's apply the \cref{crit:uuc} (\cref{eq:uuc-aligned}) to \cref{eq:update-solved} directly:
\begin{align}
    \nonumber
    1 \setsim \transpose{\pdv{\loss}{\vh_\ell}} \Delta\vh_\ell &\sim \gmult_\ell^2 \left(\norm{\vh_{\ell-1}}^2 \norm{\pdv{\loss}{\vh_\ell}}^2 + \gmult_{\ell-1}^2 \norm{\vh_{\ell-2}}^2 \norm{\transpose{\mW_\ell}\pdv{\loss}{\vh_\ell}}^2 \right) \\
    \label{eq:uucalign}
    &\sim \frac{\gmult_\ell^2}{\norm{\Delta\vh_{\ell}}^2} \left( \norm{\vh_{\ell-1}}^2 + \gmult_{\ell-1}^2 \norm{\vh_{\ell-2}}^2 \sigma_\ell^2 n_{\ell-1} \right)
\end{align}

where we understand that the passthrough term is zero for $\ell=1$. Invoking the \cref{crit:max} and the \cref{crit:ntc}:
\begin{equation}
    \label{eq:gmult-solved}
    \gmult_\ell \sim \frac{\norm{\Delta\vh_{\ell}}}{\sqrt{n_{\ell-1}}} \sim
    \begin{cases}
        1/\sqrt{n_2} & \text{if } \ell = 3\\
        \norm{\Delta\vh}/\sqrt{n_{\ell-1}} & \text{if } \ell \in {1, 2}\\
    \end{cases}.
\end{equation}

We apply \cref{eq:gmult-solved} to the other constraint relations \cref{eq:ffsize} and \cref{eq:sigma-constrained} to obtain 
\begin{equation}
    \label{eq:sigma-solved}
    \sigma_\ell \sim \frac{1}{\norm{\Delta\vh}}.
\end{equation}

\begin{SCfigure}[3][t]
  \centering
  \includegraphics[width=.55\textwidth]{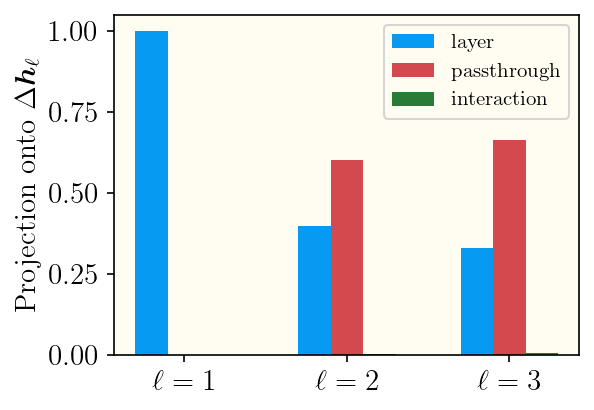}
  \caption{\textbf{Near initialization, passthrough matches layer contribution.} For a 3-layer linear models learning Gaussian data (see \cref{apdx:experiments}), we project the different update contributions onto the total representation update (see \cref{eq:update}) and measure the relative lengths. These ratios do not vary appreciably over widths and richnesses. The layer contribution is non-negligible, indicating that the \cref{crit:max} is satisfied. The interaction contribution is negligible at all layers; it may become non-negligible later in training.
  }
  \label{fig:contribs}
\end{SCfigure}

For completeness (and out of curiosity), let’s evaluate the relative size of the passthrough contributions by examining \cref{eq:uucalign}.
\begin{align}
    \nonumber
    1 &\sim \frac{\gmult_\ell^2}{\norm{\Delta\vh_{\ell}}^2} \left( \norm{\vh_{\ell-1}}^2 + \gmult_{\ell-1}^2 \norm{\vh_{\ell-2}}^2 \sigma_\ell^2 n_{\ell-1} \right) \\
    \nonumber
    &\sim 1 + \gmult_{\ell-1}^2 \norm{\vh_{\ell-2}}^2 \sigma_\ell^2 \\
    &\sim 1 + \norm{\Delta\vh_{\ell-1}} \sigma_\ell^2.
\end{align}
Plugging in, we see that the passthrough contribution for layers 2 and 3 matches the size of the layer contribution (while for layer 1, there is of course no passthrough). We demonstrate this empirically in \cref{fig:contribs}.

\cref{eq:gmult-solved} and \cref{eq:sigma-solved} deliver on the original promise: we have used all the constraints to determine the scaling of all the initial hyperparameters in terms
of a single degree of freedom, $\norm{\Delta\vh}$. Abbreviating $n_1, n_2$ as $n$, we compile the final results of the derivation in \cref{tab:results_init}.

What values can $\norm{\Delta\vh}$ take? From the error signal bound in \cref{eq:ffsize}, we know $\sigma_3^2 \lesssim  1$ which gives $\norm{\Delta\vh} \gtrsim 1$. On the other hand, we don't want the representation updates to be too large either. A reasonable criterion is $\norm{\Delta\vh_2} \lesssim \norm{\vh_2} \sim \sqrt{n}$ (see \cref{sssec:alignedgrad}). Together, these demarcate the \textit{richness scale}, which defines a continuous scale of possible choices for the richness (or activity) of training:
\begin{empheq}[box=\widefbox]{align} 
\label{eq:richnessscale}
\norm{\Delta\vh} \sim n^r \qqtext{where the richness $r$ satisfies} 0 \leq r \leq 1/2.
\end{empheq}

\begin{table}[t]
    \centering
    \begin{tabular}{lccccc}
    \toprule
    \qquad\qquad\qquad & $\gmult_\ell$ & $\sigma_\ell$ & $\norm{\vh_\ell}$ & $\norm{\gmult_\ell \Delta\mW_\ell \vh_{\ell-1}}$ & $\norm{\gmult_\ell \mW_\ell \Delta \vh_{\ell-1}}$ \\
    \midrule
    \rule{0pt}{2em}
    $\ell=1$  & $\dfrac{\norm{\Delta \vh}}{\sqrt{n_0}}$ & $\dfrac{1}{\norm{\Delta \vh}}$ & $\sqrt{n}$ & $\norm{\Delta \vh}$ & 0 \\
    \rule{0pt}{3em}
    $\ell=2$  & $\dfrac{\norm{\Delta \vh}}{\sqrt{n}}$ & $\dfrac{1}{\norm{\Delta \vh}}$ & $\sqrt{n}$ & $\norm{\Delta \vh}$ & $\norm{\Delta \vh}$ \\
    \rule{0pt}{3em}
    $\ell=3$  & $\dfrac{1}{\sqrt{n}}$ & $\dfrac{1}{\norm{\Delta \vh}}$ & $\dfrac{\sqrt{n_3}}{\norm{\Delta \vh}}$ & 1 & 1 \\[1em]
    \bottomrule
    \end{tabular}
    \caption{Choosing $\gmult_\ell$ and $\sigma_\ell$ according to this table will result in training that satisfies our three training criteria. There remains a free hyperparameter $\norm{\Delta\vh}$ which controls the richness (or activity) of training. We also report the representation sizes at initialization, as well as the layer and passthrough contributions to the representation update.}
    \label{tab:results_init}
\end{table}

\subsection{Understanding the richness scale} 
\label{ssec:understanding}

As we depict in \cref{fig:richness}, the endpoints of the richness scale correspond to well-studied training regimes. Specifically, if $r=0$ then the updates are minimal and we recover the lazy NTK training regime; on the other hand, if $r=1/2$ then the updates are maximal and we recover the feature-learning \mup\ regime. Notice that in the ``thermodynamic limit'' $n\to\infty$, we observe a discontinuous phase transition in the order parameter $\norm{\Delta\vh}/\norm{\vh}$, which vanishes everywhere along the richness scale except at \mup. We conclude that at infinite width, only \mup\ avoids the lazy regime where hidden representations evolve negligibly.

We now make several comments which explore useful intuition, address loose ends from our derivation, provide empirical evidence, and connect with other results in the literature.

\begin{figure}[t]
  \includegraphics[width=\textwidth]{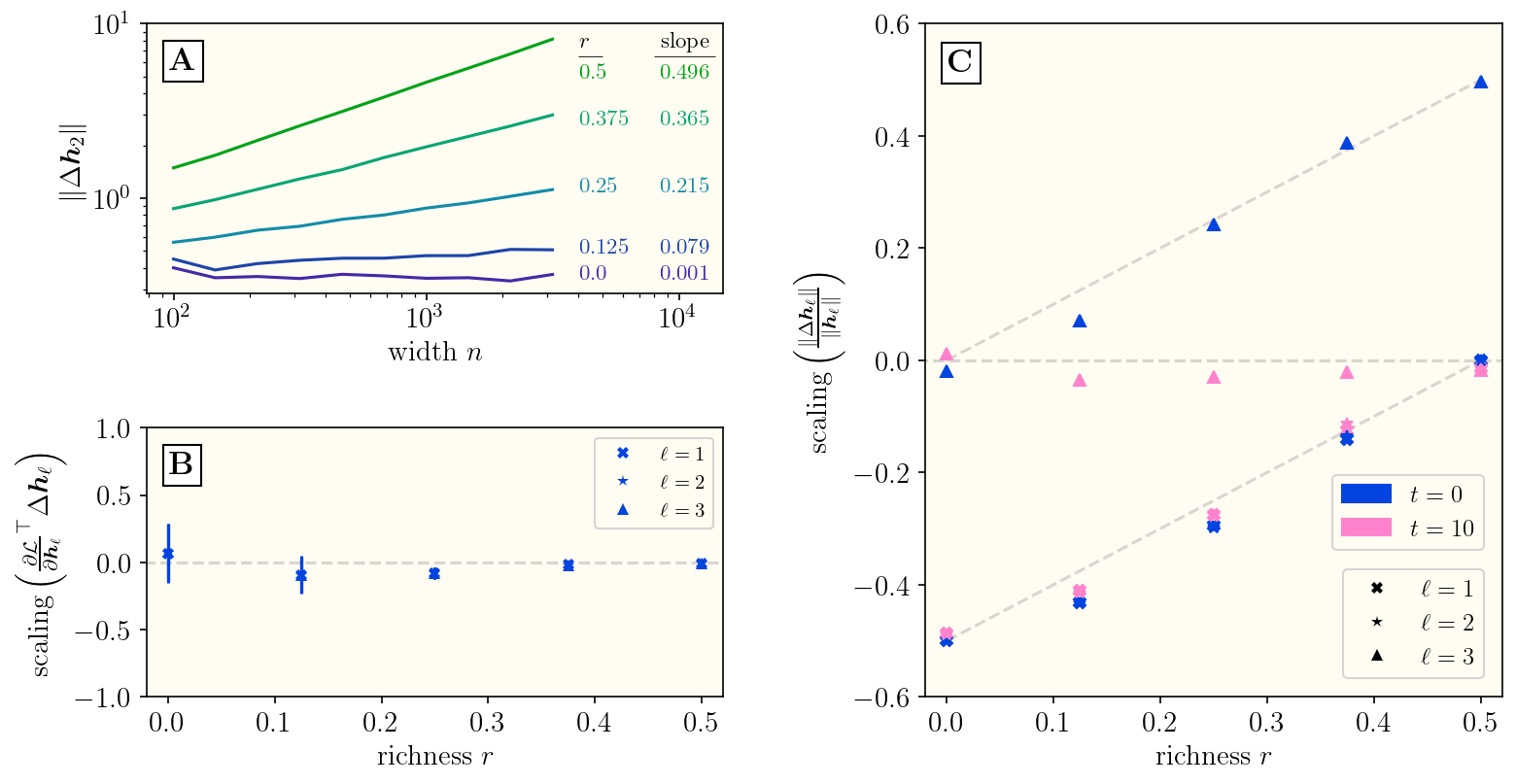}
  \caption{\textbf{Width-scaling of representations match predictions.} We report measurements of 3-layer linear models learning Gaussian data across widths and richnesses. \textbf{(A)} Depicts how we measure the scaling exponent of some scalar $s$, which we denote $\text{scaling}(s)$. In this case, $s=\norm{\Delta\vh_2}$; we measure it across widths and fit a line (on log-scale) whose slope is the measured scaling exponent. We plot the average over 20 network instances and 50 training samples. \textbf{(B)} We verify that the \cref{crit:uuc} holds at all layers and richnesses. Dotted line denotes theory prediction. \textbf{(C)} The relative sizes of representation updates match predictions. We see that the hidden representations fall on the lower dotted diagonal as predicted (low richness yields small updates). At initialization, we see the relative size of the output updates (blue triangles) fall on the upper diagonal (in the rich regime, initial outputs scale inversely with width). After a gradient step, the output size match the update size. See \cref{apdx:experiments} for details.
  }
  \label{fig:fcn-scaling}
\end{figure}

\subsubsection{Weights update to align with their input.} 
\label{sssec:weightalign}
Let's derive the weight update \cref{eq:weightupdate} using gradient descent with learning rate $\eta=1$.
\begin{align*}
    \Delta \mW_\ell^\idx{ij} &= -\pdv{\loss}{\mW_\ell^\idx{ij}} 
    = -\pdv{\loss}{\vh_\ell^\idx{k}} \pdv{\vh_\ell^\idx{k}}{\mW_\ell^\idx{ij}} \\
    &= -\pdv{\loss}{\vh_\ell^\idx{k}} \left(\delta^\idx{ik} \gmult_\ell \vh_{\ell-1}^\idx{j} \right)\\
    &= -\gmult_\ell \pdv{\loss}{\vh_\ell^\idx{i}} \vh_{\ell-1}^\idx{j}
\end{align*}
where the Kronecker delta $\delta^\idx{ik}$ indicates that the derivative $\pdv*{\vh_\ell}{\mW_\ell}$ is a sparse tensor, so its contraction with $\pdv*{\loss}{\vh_\ell}$ results in a rank-1 update. Since the input $\vh_{\ell-1}$ is the only nonzero right singular vector of the update, evidently the update is aligned. As expected, the update size is modulated by the gradient multiplier.

\begin{figure}[t]
  \includegraphics[width=\textwidth]{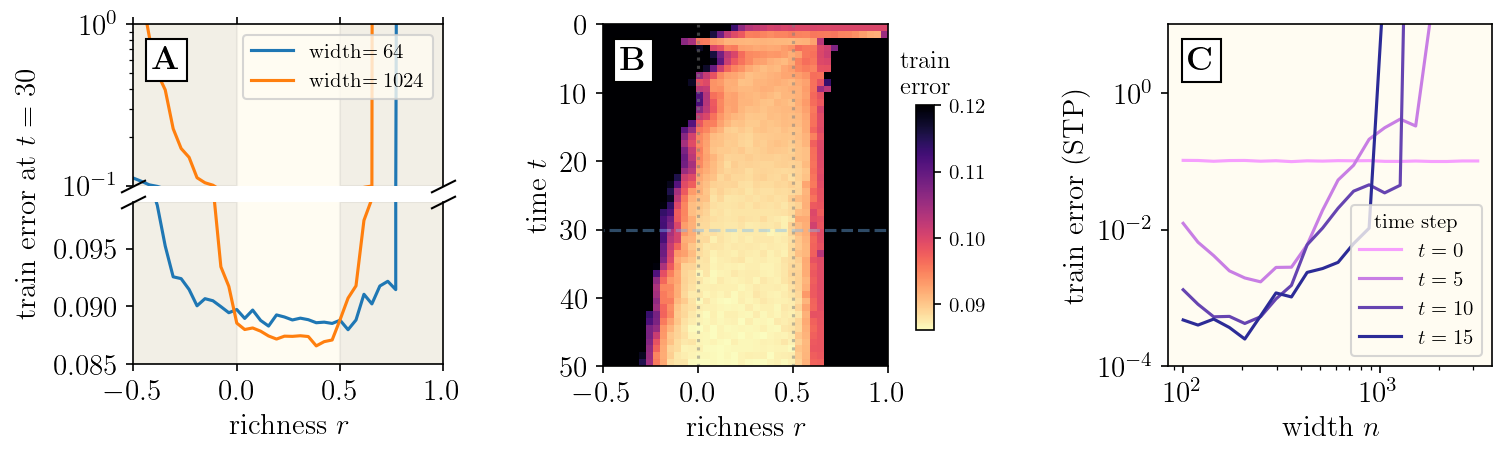}
  \caption{\textbf{Training outside the richness scale yields unstable training at large width.} We report measurements of practical convolutional network learning a minibatch of CIFAR-10. \textbf{(A)} Training is well-behaved on the richness scale ($0\leq r\leq 0.5)$; outside in the shaded regions, training error either diverges or converges very slowly. This effect becomes more prominent as width increases. \textbf{(B)} The loss dynamics of the width-1024 architecture reveals that the $r>0.5$ regime initially has reasonably-sized outputs, but training is unstable. On the other hand, in the $r<0$ regime, the initial outputs blow up. Although gradient descent eventually corrects this, the correction timescale diverges with width. The horizontal cross-section at the dashed line is the orange curve in panel A. \textbf{(C)} Here, we retain the same convolutional architecture but use standard parameterization (i.e., the default PyTorch initialization). At sufficiently large width, training diverges as predicted. However, to see this effect at practical widths, we used a global learning rate $\eta=1$ (compared to $\eta=0.1$ in the other experiments). The observed stability at moderate widths and learning rates suggests that training standard neural networks may be phenomenologically similar to training \mup\ networks. See \cref{apdx:experiments} for experimental details.
  }
  \label{fig:richness-stability}
\end{figure}

\subsubsection{Weight alignment does not magnify gradients.}
\label{sssec:alignedgrad}
We must now ask: are the conclusions of our derivation invalidated as soon as the weight matrices become correlated with the inputs? In particular, do the newly aligned weights magnify the gradients and destabilize training? It turns out that our upper bound on $\norm{\Delta\vh}$ prevents exactly this. We'll conduct a second backwards pass and examine the backprop equation for the representations:
\begin{equation*}
    \pdv{\loss}{\vh_{\ell-1}} = \gmult_\ell  \transpose{(\mW_\ell+\Delta\mW_\ell)} \pdv{\loss}{\vh_{\ell}}.
\end{equation*}
To ensure the gradient is not magnified, we need to make sure the second term doesn't dominate. We know from \cref{eq:repgradient} that the squared norm of the first term is $\gmult_\ell^2 \sigma_\ell^2 n_{\ell-1} \norm{\pdv*{\loss}{\vh_{\ell}}}^2$. The squared norm of the second term is $\gmult_\ell^4 n_{\ell-1} \norm{\pdv*{\loss}{\vh_{\ell}}}^4$. Simplifying the desired inequality, we get
\begin{equation*}
    \norm{\Delta\vh}^2 \lesssim \norm{\Delta\vh_\ell}^2 n_{\ell-1} \qqtext{for} \ell \geq 2.
\end{equation*}
Although the $\ell=2$ bound is vacuous, we see for $\ell=3$ that choosing $\norm{\Delta\vh} \lesssim \sqrt{n}$ prevents the updates from magnifying the gradient. We conclude that \mup\ achieves the largest possible representation updates; any larger, and the aligned weights would magnify the gradient signal to instability (see \cref{fig:richness-stability}).

\subsubsection{Standard parameterization yields unstable training.}
\label{sssec:stdparam}
The default initialization scheme in most deep learning frameworks sets $\gmult_\ell=1$ and $\sigma_\ell\sim\sqrt{1/n_{\ell-1}}$. This scheme does not fall anywhere on the richness scale. One can imagine it similar to \mup\, except that $\gmult_3$ and $\sigma_1$ are too large, and $\gmult_1$ is too small. This suggests that the gradient will be too large and the outputs will explode after the weights align. Practically, these shortcomings may not become evident until the network becomes sufficiently wide (see \cref{fig:richness-stability}c).

\begin{figure}[t]
  \includegraphics[width=\textwidth]{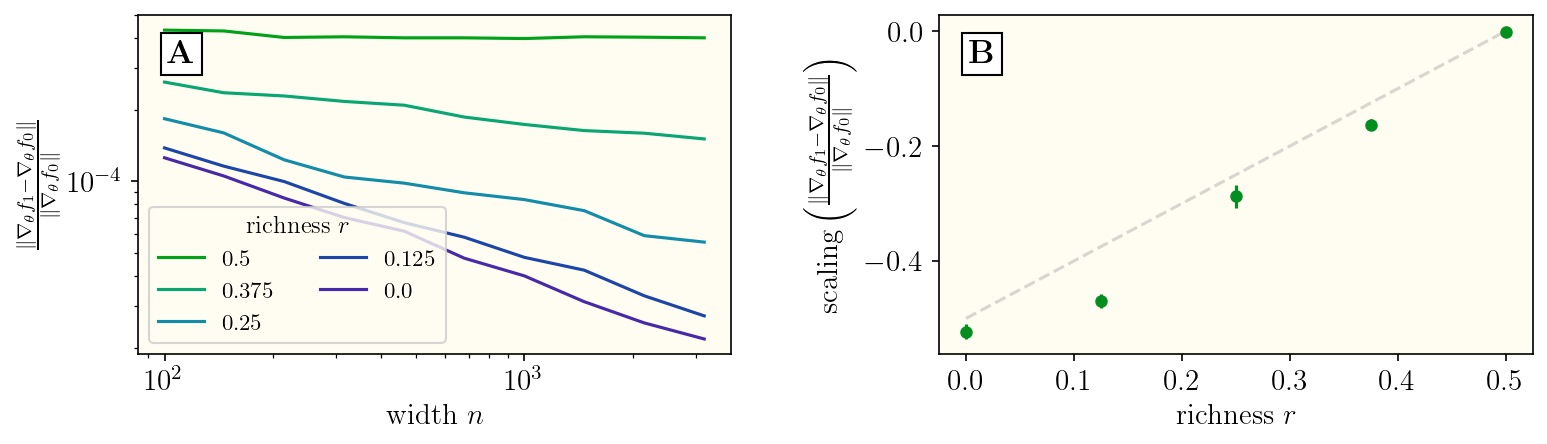}
  \caption{\textbf{Model linearization matches predictions.} For 3-layer linear models learning Gaussian data across widths and richnesses, we measure the change in the gradient across the first optimization step. \textbf{(A)} The change in the gradient decays with width in the kernel regime. (Here, the subscripts in $f_0$ and $f_1$ enumerate time steps, not layers.) \textbf{(B)} The scaling matches the prediction in \cref{eq:gradchange}.
  }
  \label{fig:linearization}
\end{figure}

\subsubsection{Models train lazily if and only if they are linearized.}
\label{sssec:lazylinearized}
Some wide networks are known to be well-approximated by their first-order Taylor approximation in parameter space:
\begin{equation*}
f(\vx; \vtheta_0+\Delta\vtheta) = f(\vx; \vtheta_0) + \transpose{\Delta\vtheta}\left(\nabla_\vtheta f(\vx;\vtheta_0)\right) + 
\underbrace{\frac{1}{2}\transpose{\Delta\vtheta}\left(\nabla^2_\vtheta f(\vx;\vtheta_0)\right)\Delta\vtheta + \cdots}_\text{negligible in the wide + lazy limit.}
\end{equation*}
Such \textit{linearized} models enjoy several nice theoretical properties, not least of which is an equivalence with (neural tangent) kernel machines. Under what conditions do wide networks become linearized kernel machines? In \cref{apdx:linearization} we show that if $\Delta\vtheta$ is a gradient descent update, the first-order gradient term scales like $\transpose{\Delta\vtheta} \left(\nabla_\vtheta f(\vx;\vtheta_0)\right) \sim 1$, while the second-order curvature term scales like
\begin{equation*}
\transpose{\Delta\vtheta}\left(\nabla^2_\vtheta f(\vx;\vtheta_0)\right) \Delta\vtheta \sim \frac{\norm{\Delta \vh}^2}{n}.
\end{equation*}
Therefore, as $n\to\infty$, the curvature term vanishes as long as $\norm{\Delta \vh}\ll \sqrt{n}$, i.e., as long as we are not in the \mup\ regime. This reflects the \textit{dynamical dichotomy theorem} of \textcite{yang2021tensor}, which states that the kernel regime and feature learning are mutually exclusive. See \cref{fig:linearization} for empirical evidence of this scaling.


\subsubsection{Small initial outputs are necessary for representation learning.}
\label{sssec:smalloutputs}
Notice that since $\norm{\vh_3} \sim 1/\norm{\Delta\vh}$, in order to achieve feature learning in the $n\to\infty$ limit, we must necessarily have small outputs at initialization. Conversely, if we want the initial outputs to be $\Theta(1)$, we must accept lazy training. Note that having small outputs at initialization is not a problem, since after the first gradient step, the readout layer will align, and the outputs will achieve the correct scale.

\subsubsection{Done correctly, model rescaling emulates training at any richness.}
\label{sssec:rescaling}
Previous works \parencite{chizat2019lazy, geiger2020disentangling, bordelon2023self} consider initializing a network in the NTK regime and toggling rich training by rescaling only the final-layer gradient multiplier and the global learning rate:
\begin{equation*}
\text{optimize}\quad \loss(\gamma^{-1} \vf^\idx{\text{NTK}}(\vx), \vy) \qtext{with learning rate} \eta = \gamma^2.
\end{equation*}
This contrasts our approach of adjusting all the gradient multipliers and initial weight scales and leaving the global learning rate architecture-independent. However, as we'll show, the two approaches are equivalent: one can exactly emulate training with richness $r$ by choosing
\begin{equation*}
\gamma \seteq n^r.
\end{equation*}
To see this, let's implement the global learning rate as gradient multipliers instead, using the procedure we describe in \cref{sssec:gradmults}. After this transformation, we have
\begin{equation*}
\eta=1 \qquad \gmult_\ell = \gamma\gmult_\ell^\idx{\text{NTK}} \qquad\sigma_\ell = \gamma^{-1}\sigma_\ell^\idx{\text{NTK}}
\end{equation*}
so that the scale of the forward pass, $\gmult_\ell\sigma_\ell$, is invariant under this transformation. Now we absorb the model rescaling coefficient into the model architecture (i.e., multiplying it into $\gmult_3$), giving us $\gmult_3=\gmult_3^\idx{\text{NTK}}$. Comparing to \cref{tab:results_init}, we see that setting $\gamma = \norm{\Delta\vh}$ allows the rescaled model to achieve training at any richness. (It's not hard to extend this argument to show that one can start with a model initialized at any richness $r_i$ and set $\gamma = n^{r-r_i}$ to achieve training with richness $r$.)

However, we now emphasize some important takeaways. First, the richness scale measures how $\norm{\Delta\vh}$ scales with the width $n$. Therefore, experiments which fix $\gamma$ (or $\gamma/\sqrt{n}$) and vary the width, as those in \textcite{geiger2020disentangling}, do not interrogate a single richness, but instead unintentionally traverse the richness scale with increasing width. Second, it is important to apply model rescaling to models initialized on the richness scale; if one tries to rescale models in the standard parameterization, as is done in \textcite{chizat2019lazy}, the model outputs will explode in the wide limit and one must resort to workarounds such as model centering or symmetrized initialization. Third, $\gamma$ cannot be chosen arbitrarily large or small; one must ensure that $1\lesssim\norm{\Delta\vh}\lesssim\sqrt{n}$. For example, as \textcite{geiger2020disentangling} report, if $\gamma \ll 1$ (i.e., $r<0$) training does not converge unless the model is centered, and if $\gamma \gg \sqrt{n}$ (i.e., $r>1/2$) then the model variance explodes. We empirically reproduce this undesired behavior in \cref{fig:richness-stability}; examining the training behavior for models off the richness scale validates our explanation.

In summary, to apply model rescaling, one should choose $r$ first and set $\gamma=n^r$, rather than treating $\gamma$ as a width-independent numerical constant. (Note that it remains interesting to ask what happens if we set $\gamma=\gamma_0 \sqrt{n}$ and vary $\gamma_0\sim 1$ as $n\to\infty$, as is done in \textcite{bordelon2023self}.)

\subsubsection{Layerwise learning rates introduce families of equivalent parameterizations.}
\label{sssec:gradmults}
Gradient descent prescribes that the weights update via a linear coupling to the gradient signal. To control the size of the update, one can either change the coupling strength (i.e., the learning rate) or the size of the gradient signal (via the $\gmult_\ell$). It's not hard to show that this is a \textit{gauge symmetry}, i.e., both approaches yield the same training dynamics. Consider freezing all layers prior to $\ell$ and computing the representation update with a layerwise learning rate $\eta_\ell$:
\begin{equation*}
    \Delta \vh_\ell = \gmult_\ell\Delta\mW_\ell\vh_{\ell-1}
    = -\eta_\ell \gmult_\ell^2 \norm{\vh_{\ell-1}}^2 \pdv{\loss}{\vh_\ell}.
\end{equation*}
Clearly, we can achieve the same learning trajectory as before (where $\eta_\ell=1$ and $\gmult_\ell\equiv \bar g$) by simply setting $\eta_\ell=\bar g^2$ and $\gmult_\ell=1$. (To keep the initial feedforward activations the same, we also need to multiply $\sigma_\ell$ by $\bar g$.) One can easily treat the general case with no frozen layers by induction. So, we don't lose anything by choosing to use gradient multipliers and simply setting $\eta=1$.

\begin{SCfigure}[3][t]
  \includegraphics[width=0.68\textwidth]{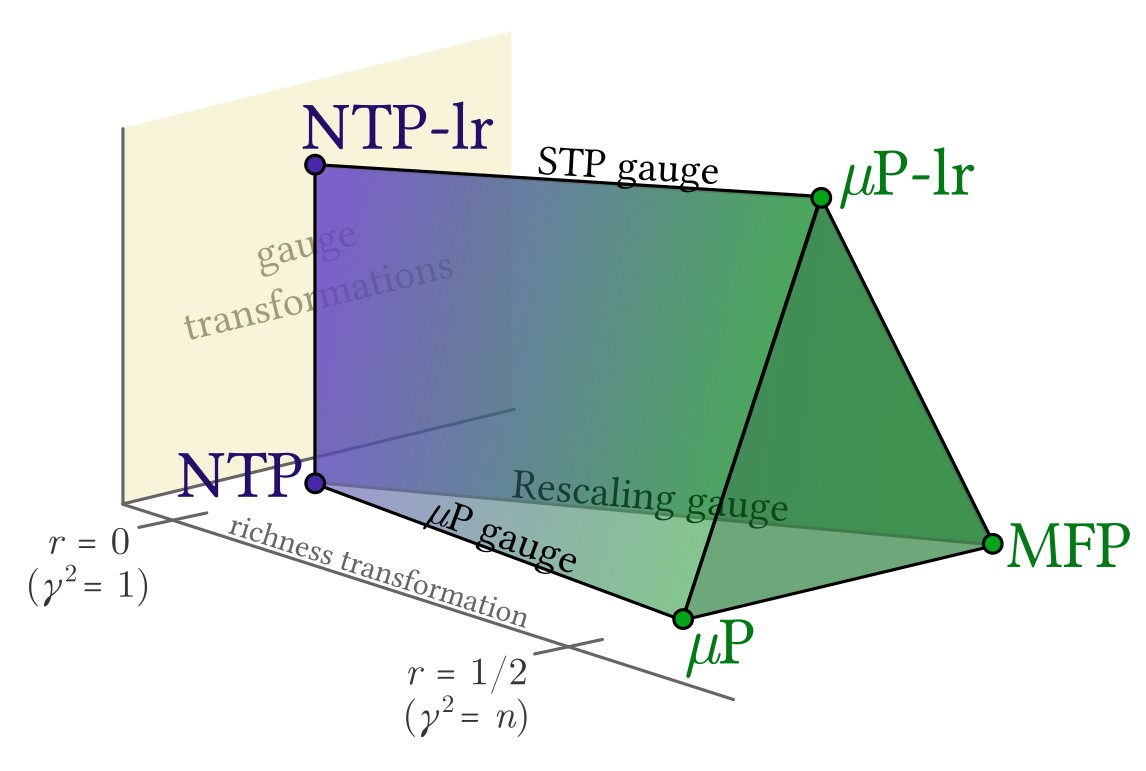}
  \caption{\textbf{Gauge freedom yields families of equivalent parameterizations.} We depict the space of parameterizations, indicating five typical parameterizations and the gauges connecting lazy ones to rich ones. Each gauge offers its own implementation of the richness scale. Gauge transformations (e.g., \mup$\to$MFP) don't affect training dynamics.
  }
  \label{fig:gauges}
\end{SCfigure}

Introducing layerwise learning rates yields redundant degrees of freedom, which collectively constitute a \textit{gauge freedom} in infinite-width parameterizations. This gauge freedom is responsible for the multitude of parameterizations discussed in the literature. We summarize the three most common gauges in \cref{fig:gauges} and \cref{tab:parameterizations}. One obtains these gauges by choosing one of $\eta_\ell=1$, $\sigma_\ell=1$, or $\gmult_\ell=1$, for all $\ell$. The choice of gauge does not affect observables; only the richness hyperparameter $r$ (or $\gamma\equiv n^r$) affects training dynamics.

In this tutorial, we chose the $\eta_\ell=1$ gauge, which we call the \mup\ gauge. In \cref{sssec:rescaling}, we showed how to use model rescaling with $\sigma_\ell=1$ to achieve the same richness scale; we call this the rescaling gauge, and the rich limit is known as ``mean-field parameterization'' in the literature \parencite{mei2019mean, bordelon2023self}. Setting $\gmult_\ell=1$ instead, one obtains what we call STP gauge \parencite{everett2024scaling, yang2023spectral}, since, like standard parameterization, there are no gradient multipliers in the architecture. However, we emphasize that STP gauge does \textit{not} contain standard parameterization, which does not lie on the richness scale (see \cref{sssec:stdparam}). Code to implement these parameterizations can be found at \url{https://github.com/dkarkada/mupify}.

\bgroup
\setlength\tabcolsep{20pt}
\begin{table}[t]
    \centering
    \begin{tabular}{c|c|c|c}
    \toprule
    & \mup\ gauge & Rescaling gauge & STP gauge\\
    \midrule
    \rule{0pt}{2em}
    $\eta_\ell$  & $1$ & $\gamma^2$ & $\dfrac{\gamma^2}{d} \quad \dfrac{\gamma^2}{n} \quad \dfrac{1}{n}$\\
    \rule{0pt}{3em}
    $\sigma_\ell^2$ & $\dfrac{1}{\gamma^2}$ & $1$ & $\dfrac{1}{d} \quad \dfrac{1}{n} \quad \dfrac{1}{\gamma^2 n}$ \\
    \rule{0pt}{3em}
    $\gmult_\ell^2$ & $\dfrac{\gamma^2}{d} \quad \dfrac{\gamma^2}{n} \quad \dfrac{1}{n}$ & $\dfrac{1}{d} \quad \dfrac{1}{n} \quad \dfrac{1}{\gamma^2 n}$ & $1$\\[1em]
    \bottomrule
    \end{tabular}
    \caption{Layerwise learning rates introduce a gauge freedom (i.e., equivalence relation) to the space of wide parameterizations. Changing gauge does not impact training dynamics. We summarize three gauges which appear in the literature. The \mup\ gauge is the gauge derived in this tutorial (c.f. \cref{tab:results_init}, where we used $\norm{\Delta\vh}$ to denote the activity hyperparameter $\gamma$). The rescaling gauge was described in \cref{sssec:rescaling}; the feature-learning parameterization associated with this gauge is often referred to as ``mean field parameterization.'' The STP gauge is akin to standard parameterization in that there are no gradient multipliers in the architecture; however, to satisfy our training criteria we must use layerwise learning rates and change the scale of the readout weights (see \cref{sssec:stdparam}).} 
    \label{tab:parameterizations}
\end{table}
\egroup

\subsubsection{Our conclusions empirically hold for practical architectures.}
\label{sssec:realisticmodels}
Our toy model is unrealistic on several counts. We train on a single sample, whereas minibatch SGD is standard; we neglect nonlinear activation functions; and we use a fully-connected network despite its inferior inductive bias (compared to, e.g., convolutional networks for vision tasks). In \cref{fig:richness-stability} and \cref{fig:cnn-scaling} we present empirical evidence that our conclusions hold for realistic training setups. In particular, we train a Myrtle-5 CNN (depth-5, ReLU activations, global average pooling, SGD+momentum with minibatch size 32; described in \cite{shankar2020neural}) to learn CIFAR10.
\section{Conclusions}
\label{sec:conclusion}

We've shown how to use scaling arguments to derive the \textit{richness scale} for training behavior; we've discussed several consequences and related perspectives; and, we've provided empirical evidence supporting our nonrigorous derivations. We conclude with a few final comments.

The infinite-width NTK regime is well-studied, and many questions regarding convergence and generalization have been answered. However, understanding the emergence of feature learning remains an open problem. Prior works perturbatively computed finite-width corrections to the NTK and conjectured that these corrections were the source of feature learning in practical networks. We now know that \textit{richness} is another distinct mechanism for neural networks to achieve feature learning. As we've seen, richer architectures have larger gradient multipliers, so rich training is likely related to the empirical phenomena associated with large learning rates. Disentangling these effects is a ripe area for future research.

A core idea in our derivation is that the representations $\vh_\ell$ play a key role. It's not enough to study the weight matrices alone; in the end, it's the hidden representations that form the bridge between the input data and the desired output. In our simple derivation, we only controlled the \textit{size} of $\vh_\ell$ and $\Delta\vh_\ell$, but to better understand the rich regime, it will be important to assess and control the \textit{quality} of the learned representations.

A primary difficulty in characterizing the learned representations is that the structure of natural data distributions is poorly understood. To see why understanding this structure is necessary, consider that learned representations typically improve generalization \parencite{lee2020finite}, and characterizing generalization performance typically requires ``omniscient'' knowledge of the interaction between the data distribution and the model's inductive bias \parencite{bordelon2020spectrum, simon2023more}. Studying the infinite-width rich regime may shed light on this data-model interaction in practical neural networks, thereby advancing the science of deep learning.

\textbf{Acknowledgements.}
I am grateful to Jamie Simon for inspiration, guidance, and helpful comments; Libin Zhu, Jonathan Shi, and Michael DeWeese for useful discussions; and Jared Hrebenar for love and encouragement. This work is supported by the UC Berkeley Department of Physics.

\printbibliography[title={References}]

\newpage
\appendix
\section{Linearization in wide linear models}
\label{apdx:linearization}

Infinitely-wide networks trained in the lazy regime are known to be equivalent to kernelized linear models, i.e., models of the form $\vf(\vx;\vw)=\transpose{\vw}\phi(\vx)$ for some kernel mapping $\phi(\cdot)$. For a neural network $\vf(\vx;\vtheta)$ with model parameters $\vtheta$, this happens when the network is \textit{linearized}, i.e., when the second (and higher) order terms of the Taylor expansion of $\vf(\vx;\vtheta_0+\Delta\vtheta)$ in $\Delta\vtheta$ become negligible:
\begin{equation}
\label{eq:linearizednet}
\vf(\vx; \vtheta_0+\Delta\vtheta) - \vf(\vx; \vtheta_0) \approx \transpose{\Delta\vtheta }\nabla_\vtheta\vf(\vx;\vtheta_0).
\end{equation}
Here, we see the residual is a kernelized linear model with $\phi(\vx)= \nabla_\vtheta\vf(\vx;\vtheta_0)$ being the kernel mapping. Our goal here is to show that \cref{eq:linearizednet} follows from lazy training in the infinite-width limit. To do this, we will show that lazy training causes the curvature of the model in parameter space to vanish with respect to the size of the gradient:
\begin{equation*}
\transpose{\Delta\vtheta} \mH \Delta\vtheta \ll \transpose{\Delta\vtheta }\nabla_\vtheta\vf(\vx;\vtheta_0)
\quad\qtext{where} \mH \defn \nabla^2_\vtheta \vf(\vx;\vtheta_0) \ \text{is the Hessian matrix.}
\end{equation*}
Under this condition, the model reduces to its first-order Taylor approximation, and so our learning problem reduces to kernel regression with the aforementioned kernel mapping. It is important to note that the sufficient condition for kernel learning is linearizing $\vf(\vx;\vtheta)$ \textit{in parameter space}; in general, it may still be nonlinear in the inputs. However, for convenience in our present analysis, we will \textit{also} assume that our model is linear in our inputs (as we did previously). To disambiguate going forward, we will use \textit{linearized} to refer to models that have become linear in parameter space due to the lazy infinite-width limit, and \textit{linear} to refer to models which are linear in the inputs due to lacking nonlinear activation functions.

We will consider a depth-$L$ scalar linear model
\begin{equation*}
f(x) = \gmult_L \transpose{\vw_L} \left(\prod_{1<\ell<L} \gmult_\ell\mW_\ell\right) \gmult_1 \vw_1 x
\end{equation*}
on a scalar input $x$. We define the hidden representations $\vh_\ell$ as before. Again we take a wide limit, with all hidden representations having equal width $n$. We must consider deeper models ($L>3$) because, as we will see, the scaling of the curvature term depends primarily on the coupling between square layers, so having a single square layer is insufficient for treating the general case. However, we will always assume $L\sim 1$ w.r.t. to the wide network limit.

For convenience, we will analyze the curvature term for the first gradient step only, so that $\Delta\vtheta 
\sim \nabla_\vtheta f(\vx;\vtheta_0)$. If we assume that training converges at a width-independent rate, then we may assume that a lack of curvature in the first step will imply flatness throughout training.

Using the \cref{crit:uuc} and our solution for $\gmult_\ell$ for intermediate layers, we know that the gradient w.r.t. an intermediate layer parameter has size
\begin{equation*}
\abs{\pdv{f}{\mW_\ell^\idx{ij}}} = \abs{\pdv{f}{\vh_\ell^\idx{i}} \gmult_\ell \vh_{\ell-1}^\idx{j}} \sim \frac{\gmult_\ell}{\sqrt{n}\norm{\Delta \vh}} \sim \frac{1}{n}.
\end{equation*}
It is easy to show that the readin and readout derivatives are not too large, so we may estimate the gradient term by considering the $\Theta(n^2)$ intermediate-layer parameters:
\begin{equation*}
\transpose{\Delta\vtheta }\nabla_\vtheta\vf(\vx;\vtheta_0) \sim \norm{\nabla_\vtheta\vf(\vx;\vtheta_0)}^2 \sim \sqrt{n^2 (1/n)^2} \sim 1.
\end{equation*}
Let us write the Hessian for our deep linear model. In general, the two Hessian indices refer to parameters at possibly different layers, $\ell$ and $\ell'$. Invoking the symmetry of the Hessian, let us choose $\ell \geq \ell'$. Then
\begin{equation*}
\pdv{f}{\mW_\ell^\idx{ij}}{\mW_{\ell'}^\idx{pq}} = \gmult_\ell \pdv{f}{\vh_\ell^\idx{i}} \pdv{\vh_{\ell-1}^\idx{j}}{\mW_{\ell'}^\idx{pq}} = \gmult_\ell \pdv{f}{\vh_\ell^\idx{i}} \pdv{\vh_{\ell-1}^\idx{j}}{\vh_{\ell'}^\idx{p}} \gmult_{\ell'} \vh_{\ell'-1}^\idx{q}
\end{equation*}
We will find that the scale of this Hessian element depends sensitively on the ``layer connection derivative:''
\begin{equation*}
\pdv{\vh_{\ell-1}^\idx{j}}{\vh_{\ell'}^\idx{p}} \sim 
    \begin{cases}
        0 & \text{if } \ell = \ell'\\
        \delta^\idx{jp} & \text{if } \ell - 1 = \ell' \\
        1/\sqrt{n_{\ell'-1}} & \text{otherwise}
    \end{cases}
\end{equation*}
where $\delta^\idx{\cdot \cdot}$ is a Kronecker delta. The first case tells us that the Hessian element of two parameters belonging to the same layer vanishes; this follows from the linearity of the model. The second case tells us that the Hessian element of two parameters from adjacent layers vanishes unless they are connected to the same neuron. The final case can be verified by direct computation, and follows from our feedforward constraint \cref{eq:ffsize}.

We are now ready to compute the curvature term for one gradient step:
\begin{align*} 
\transpose{\Delta\vtheta} \mH \Delta\vtheta
&\sim 
\sum_{\ell, \ell'} \pdv{f}{\mW_\ell^\idx{ij}}\pdv{f}{\mW_\ell^\idx{ij}}{\mW_{\ell'}^\idx{pq}} \pdv{f}{\mW_{\ell'}^\idx{pq}} \\ 
&\sim
\sum_{\ell, \ell'} \left(\gmult_\ell \pdv{f}{\vh_\ell^\idx{i}} \vh_{\ell-1}^\idx{j}\right)
\left(
\gmult_\ell \pdv{f}{\vh_\ell^\idx{i}} \pdv{\vh_{\ell-1}^\idx{j}}{\vh_{\ell'}^\idx{p}} \gmult_{\ell'} \vh_{\ell'-1}^\idx{q} \right)
\left(\gmult_{\ell'} \pdv{f}{\vh_{\ell'}^\idx{p}} \vh_{{\ell'-1}}^\idx{q}\right) \\
&\sim
\sum_{\ell, \ell'} \left(\gmult_\ell^2 \gmult_{\ell'}^2 \norm{\pdv{f}{\vh_\ell}}^2 \norm{\vh_{{\ell'-1}}}^2 \right) \left(\vh_{\ell-1}^\idx{j} \pdv{\vh_{\ell-1}^\idx{j}}{\vh_{\ell'}^\idx{p}} \pdv{f}{\vh_{\ell'}^\idx{p}}\right)
\end{align*}
after contracting indices $i$ and $q$. Although the sum is over all layer pairs $\ell, \ell'$, the reader can verify that the terms coming from the readin and readout layers do not dominate, so we may hereforth assume that $\ell$ and $\ell'$ index intermediate layers. In this case, $n_\ell=n_{\ell'}=n_{\ell'-1}\equiv n$, so the first parenthesized factor in the last line simplifies and we obtain
\begin{equation*} 
\transpose{\Delta\vtheta} \mH \Delta\vtheta
\sim 
\sum_{\ell, \ell'} \frac{\norm{\Delta \vh}^2}{n} \left(\vh_{\ell-1}^\idx{j} \pdv{\vh_{\ell-1}^\idx{j}}{\vh_{\ell'}^\idx{p}} \pdv{f}{\vh_{\ell'}^\idx{p}}\right).
\end{equation*}
We now evaluate the remaining factor according to the ``layer connection derivative'' above:
\begin{equation*}
\vh_{\ell-1}^\idx{j} \pdv{\vh_{\ell-1}^\idx{j}}{\vh_{\ell'}^\idx{p}} \pdv{f}{\vh_{\ell'}^\idx{p}} \sim 
    \begin{cases}
        0 & \text{if } \ell = \ell'\\
        \transpose{\vh_{\ell'}} (\pdv*{f}{\vh_{\ell'}}) \sim 1 & \text{if } \ell - 1 = \ell' \text{ (by \cref{crit:uuc}) } \\
        1/\norm{\Delta\vh} & \text{otherwise (by \cref{crit:uuc})}
    \end{cases}
\end{equation*}
Plugging in, we finally obtain
\begin{equation*} 
\transpose{\Delta\vtheta} \mH \Delta\vtheta
\sim 
\frac{\norm{\Delta \vh}^2}{n} \left(L + \frac{L^2}{\norm{\Delta\vh}}\right) \sim 
\frac{\norm{\Delta \vh}^2}{n},
\end{equation*}
where in the last scaling we used the fact that $L\sim1$ and $\norm{\Delta \vh}\gtrsim 1$.\footnote{It is interesting to note that curvature in the finite-width NTK regime receives equal contributions from all pairs of layers, whereas curvature in the \mup\ regime is dominated by parameter pairs from adjacent layers.} Comparing this curvature term to our gradient term, we see that the model is linearized if 
\begin{equation*}
\lim_{n \to\infty}
\frac{\norm{\Delta \vh}^2}{n} \ll 1.
\end{equation*}
The only point on the richness scale which avoids this condition is \mup, $\norm{\Delta \vh}\sim\sqrt{n}$. Infinitely wide models initialized anywhere else train in the kernel regime.

\textbf{Change in gradient.} It's experimentally easier to simply store the gradient and track how it changes over optimization. The change in the gradient over a single optimization step scales like
\begin{equation*}
\norm{\nabla_\vtheta f(\vx;\vtheta_1) - \nabla_\vtheta f(\vx;\vtheta_0)}
\sim \norm{\mH \nabla_\vtheta f(\vx;\vtheta_0)} \sim \norm{\mH \Delta\vtheta}.
\end{equation*}
Bootstrapping the previous calculation, we see that the typical element of this gradient change vector scales like
\begin{align*} 
\mH^\idx{ij,pq} \Delta\vtheta^\idx{pq}
&\sim
\sum_{\ell, \ell'} 
\left(\gmult_\ell \pdv{f}{\vh_\ell^\idx{i}} \pdv{\vh_{\ell-1}^\idx{j}}{\vh_{\ell'}^\idx{p}} \gmult_{\ell'} \vh_{\ell'-1}^\idx{q} \right)
\left(\gmult_{\ell'} \pdv{f}{\vh_{\ell'}^\idx{p}} \vh_{{\ell'-1}}^\idx{q}\right) \\
&\sim
\sum_{\ell, \ell'} \left(\gmult_\ell \gmult_{\ell'}^2 \pdv{f}{\vh_\ell^\idx{i}} \norm{\vh_{{\ell'-1}}}^2 \right) \left( \pdv{\vh_{\ell-1}^\idx{j}}{\vh_{\ell'}^\idx{p}} \pdv{f}{\vh_{\ell'}^\idx{p}}\right) \\
&\sim
\sum_{\ell, \ell'} 
\left( \frac{\norm{\Delta \vh}^3}{n^{3/2}} \pdv{f}{\vh_{\ell}^\idx{i}} n\right)
\left( 
\delta^\idx{\ell',\ell-1}\pdv{f}{\vh_{\ell'}^\idx{j}} + \pdv{f}{\vh_{\ell'}^\idx{j}} \right)
\end{align*}
where again we restrict our attention to the intermediate layers so that $n_\ell=n_{\ell'}=n_{\ell'-1}\equiv n$. Simplifying, we find
\begin{equation}
\label{eq:gradchange}
\norm{\mH \Delta\vtheta} \sim \frac{\norm{\Delta \vh}^2}{n}\left(\frac{L}{\norm{\Delta \vh}\sqrt{n}}+\frac{L^2}{\norm{\Delta \vh}\sqrt{n}}\right)\sqrt{n^2} \sim \frac{\norm{\Delta \vh}}{\sqrt{n}}.
\end{equation}
The first two factors account for the scaling of a single element (above), and the final factor accounts for the norm of the gradient change vector, whose dimension scales like $n^2$. This is the scaling for the gradient change we observe in \cref{fig:linearization}.
\newpage
\section{Experimental details}
\label{apdx:experiments}

We perform scaling experiments on two learning tasks, each across a variety of widths and richnesses, using gradient descent with global learning rate $\eta=0.1$ and momentum $\beta=0.9$. The \textit{linear task} is training the 3-layer fully-connected linear model described in the text to learn random Gaussian data ($n_0=n_3=10$). The \textit{practical task} is training the Myrtle-5 convolutional architecture described in \cite{shankar2020neural} to learn CIFAR-10:
\begin{lstlisting}[language=Python]
    def forward(self, x):
        h1 = F.relu(self.conv1(x))
        h2 = self.pool(F.relu(self.conv2(h1)))
        h3 = self.pool(F.relu(self.conv3(h2)))
        h4 = self.gap(F.relu(self.conv4(h3)))
        assert h4.shape[1:] == (width, 1, 1)
        h5 = self.readout(h4.squeeze())
        return h5
\end{lstlisting}
where the convolutions are all $3\times 3$ with ``same'' padding so that the conv layers do not change the spatial shape, and we exclusively use average pooling ($2\times 2$ for all pooling layers except the final global-average-pool, which is $8\times 8$). In both learning tasks, all conv layers and linear layers are custom implementations using the gradient multipliers and initialization scale previously derived. The width of a convolutional feature map is simply the number of channels (which is generally \textit{not} the dimension of the flattened feature vector). We train all models using PyTorch.

To estimate scaling exponents, we simply fit a line to the log-scaled data. We sometimes restrict the fit to medium/large widths to mitigate unwanted finite-size effects. Errorbars in scaling plots report the best-fit errors for the slope. It is important to note that the error bars do not reflect statistical variation over multiple trials, nor do they reflect variations due to finite-width effects.

\begin{figure}[t]
  \includegraphics[width=\textwidth]{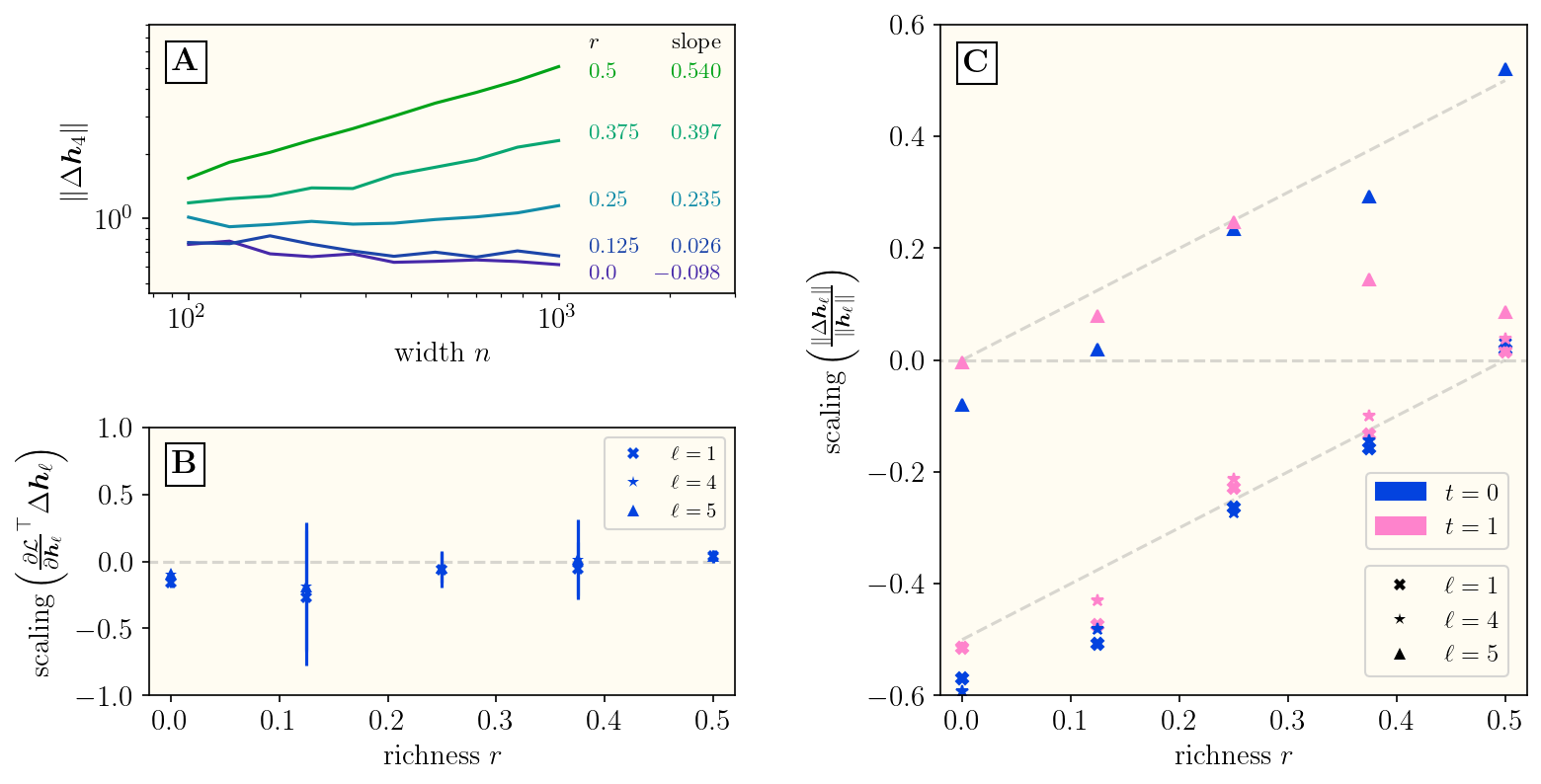}
  \caption{\textbf{Width-scaling of representations match predictions.} We repeat the scaling experiment on the practical task. The predictions are validated, apart from an apparent theory/experiment discrepancy for the relative size of the output update after weight alignment (i.e., pink triangles stray from horizontal line). We discuss the origin of this apparent discrepancy on the previous page.
  }
  \label{fig:cnn-scaling}
\end{figure}

The linear model scaling experiment (\cref{fig:fcn-scaling} and \cref{fig:contribs}) and the linearization experiment (\cref{fig:linearization}) perform the linear task; the former uses minibatch size 1, and the latter uses minibatch size 256. The training stability experiment (\cref{fig:richness-stability}) and the CNN scaling experiment (\cref{fig:cnn-scaling}) perform the practical task; the former uses minibatch size 128, and the latter uses minibatch size 32. None of our main results are sensitive to the choice of minibatch size.

The CNN scaling experiment (\cref{fig:cnn-scaling}) exhibits an apparent theory/experiment discrepancy for the relative size of the output update after weight alignment. However, this is simply an artifact of the smaller learning rate. Theory predicts: (a) output updates should never exhibit any width-dependence at any richness (c.f. the \cref{crit:ntc}), and (b) after alignment, the outputs $\vh_{L,(t=1)}=\vh_{L,(t=0)}+\Delta\vh_{L,(t=0)}$ should be the same size as the labels (i.e., width-independent). If both of these are satisfied, then after a few optimization steps the relative output update should not scale with width. However, we empirically see that at $t=1$ there is still mild scaling at intermediate richness. The reason for this seeming discrepancy is that prediction (b) is not satisfied after only 1 gradient step; the global learning rate is slightly too small, and the first output update $\Delta\vh_{L,(t=0)}$ is not large enough at intermediate richness to ``overwrite'' the width-dependence of $\vh_{L,(t=0)}$. Contrast this with the behavior near $r=0$, where there is no width-dependence to begin with, and the behavior near $r=1/2$, where the initial $\vh_{L,(t=0)}$ are small enough that they are dominated by the first update $\Delta\vh_{L,(t=0)}$. In summary, this discrepancy may be eliminated by using a slightly larger learning rate (i.e., $\eta=0.5$); the \cref{crit:ntc} is always satisfied; and, the experiments are consistent with our derivation.

\newpage

\end{document}